 \documentclass[pmlr,twocolumn,10pt]{jmlr} % W&CP article

\usepackage{booktabs}

\makeatletter
\renewcommand\paragraph{\@startsection{paragraph}{4}{\z@}%
   {0.5ex \@plus 0.2ex \@minus 0.2ex}% space above
   {-0.5em}% space after heading (negative makes it run-in)
   {\normalfont\normalsize\bfseries}}
\makeatother
\makeatletter
\renewcommand\section{\@startsection{section}{1}{\z@}%
  {0.8ex \@plus 0.2ex \@minus 0.2ex}% space above
  {0.6ex}% space below
  {\normalfont\large\bfseries}}

\renewcommand\subsection{\@startsection{subsection}{2}{\z@}%
  {0.6ex \@plus 0.2ex \@minus 0.2ex}% space above
  {0.4ex}% space below
  {\normalfont\normalsize\bfseries}}

\renewcommand\paragraph{\@startsection{paragraph}{4}{\z@}%
  {0.4ex \@plus 0.2ex \@minus 0.2ex}% space above
  {-0.5em}% space after heading
  {\normalfont\normalsize\bfseries}}
\makeatother
\usepackage{etoolbox}
\makeatletter
\patchcmd{\@afterheading}{\@afterindenttrue}{\@afterindentfalse}{}{}
\makeatother
\newcommand{\lead}[1]{\noindent\textbf{#1}}

\author{%
\Name{Franklin Lee} \Email{franklin.lee@stonybrook.edu}\\
\addr Jericho Senior High School, USA
\AND
\Name{Tengfei Ma} \Email{tengfei.ma@stonybrookmedicine.edu}\\
\addr Stony Brook University, USA
}

\usepackage{siunitx}

\usepackage[switch]{lineno}
\usepackage{float}
\usepackage{ragged2e}

\theorembodyfont{\upshape}
\theoremheaderfont{\scshape}
\theorempostheader{:}
\theoremsep{\newline}

\jmlrworkshop{Machine Learning for Health (ML4H) 2025} % W&CP title

 \title[Short Title]{Dual-Pathway Fusion of EHRs and Knowledge Graphs for Predicting Unseen Drug-Drug Interactions}

\begin{document}

\maketitle

\begin{abstract}
% \tm{this title is not good enough. Probably focus more on unseen, and the addition of EHRs. I got two suggestions from Gemini, "Predicting Unseen Drug-Drug Interactions: A Dual-Pathway Model Fusing EHRs and Knowledge Graphs" or "DualTrack-DDI: Leveraging EHRs for Unseen Drug-Drug Interaction Prediction via Knowledge Distillation". Either one is fine. }
Drug–drug interactions (DDIs) remain a major source of preventable harm, and many clinically important mechanisms are still unknown. Existing models either rely on pharmacologic knowledge graphs (KGs), which fail on unseen drugs, or on electronic health records (EHRs), which are noisy, temporal, and site-dependent. We introduce, to our knowledge, the first system that conditions KG relation scoring on patient-level EHR context and distills that reasoning into an EHR-only model for zero-shot inference. A fusion “Teacher” learns mechanism-specific relations for drug pairs represented in both sources, while a distilled “Student” generalizes to new or rarely used drugs without KG access at inference. Both operate under a shared ontology (set) of pharmacologic mechanisms (drug relations) to produce interpretable, auditable alerts rather than opaque risk scores. Trained on a multi-institution EHR corpus paired with a curated DrugBank DDI graph, and evaluated using a a clinically aligned, decision-focused protocol with leakage-safe negatives that avoid artificially easy pairs, the system maintains precision across multi-institutuion test data, produces mechanism-specific, clinically consistent predictions, reduces false alerts (higher precision) at comparable overall detection performance (F1), and misses fewer true interactions compared to prior methods. Case studies further show zero-shot identification of clinically recognized CYP-mediated and pharmacodynamic mechanisms for drugs absent from the KG, supporting real-world use in clinical decision support and pharmacovigilance.
\end{abstract}


\paragraph*{Data and Code Availability}
We use a DrugBank knowledge graph with $R{=}86$ relation types over $1{,}710$ drugs. Per-drug EHR embeddings are computed from $\sim$200M de-identified TriNetX (learn more about TriNetX at trinetx.com) events via PyHealth. TriNetX provides harmonized, multi-institution EHR data (standard code sets such as ICD-10, RxNorm, LOINC; routinely refreshed) across 220+ healthcare organizations in 30 countries. Experiments restrict to drugs in both sources. For a candidate pair $(h, t)$, we form pairwise EHR features $x{=}x_{\text{EHR}}(h, t)$ shared by all EHR baselines and the student \citep{yang2023pyhealth, wishart2018drugbank, trinetx, Wang2024accurateandinterpretable}.
The code is available at https://tinyurl.com/mu3tuc6w. Additional information about TriNetX is in Appendix~\ref{app:trinetx}.
% This initial paragraph is \textbf{mandatory}. Briefly state what data you
% use (including citations if appropriate) and whether and where the data are
% available to other researchers.
% If you are not sharing code, you must explicitly state that you are not
% making your code available. If you are making your code available, then
% at the time of submission for review, please include your code as
% supplemental material or as a code repository link; in either case, your
% code must be anonymized. If your paper is accepted, then you should
% de-anonymize your code for the camera-ready version of the paper. \emph{If
% you do not include this data and code availability statement for your
% paper, or you provide code that is not anonymized at the time of
% submission, then your paper will be desk-rejected.} Your experiments later
% could refer to this initial data and code availability statement if it is
% helpful (e.g., to avoid restating what data you use).

% Remove extra space between paragraphs everywhere
% (If you load \usepackage{parskip}, remove it.)
\setlength{\parskip}{0pt}
\setlength{\parindent}{1.5em} % restore normal indent

\paragraph*{Institutional Review Board (IRB)} Our de-identified data does not require Institutional Review Board approval.

\section{Introduction}
\label{sec:intro}

Drug-drug interactions (DDIs) can cause serious adverse reactions and pose a significant challenge to healthcare. In recent years, many unknown DDIs are emerging and predicting potentially new drug interactions has attracted increasing attention \citep{wang2024zeroddizeroshotdrugdruginteraction,Wang2024accurateandinterpretable}. At the bedside, pharmacists need precision-first, mechanism-specific alerts (e.g., “CYP inhibition”) that reflect the current patient context \citep{shang2019gamenet,yang2021safedrug} and remain effective for new or rarely used drugs lacking prior evidence \citep{huang2024txgnn,wang2023zerobind}.

Previous computational approaches often rely on existing DDI knowledge graphs (KGs) and transform DDI prediction into a link-prediction problem \citep{zitnik2018decagon}. KG link prediction interpolates among seen drugs; however, such models struggle in zero-shot settings when a target drug lacks KG edges \citep{huang2024txgnn,wang2023zerobind}. Text-only zero-shot methods infer interactions from curated drug descriptions (e.g., TextDDI) but depend on the availability and quality of those texts (performance can degrade when descriptions are missing or noisy) limiting robustness in practice \citep{Zhu_2023}. Researchers have used electronic health record (EHR) data, a rich source for recording drug side effects, to screen DDIs; however, signals mined from EHRs are noisy, temporal, and cohort-dependent, and often transfer poorly across sites and timeframes \citep{choi2016retain,wu2021ddiwas}. More fundamentally, a static–dynamic mismatch separates population-level, timeless KG relations from patient-specific, time-varying EHR signals; reasoning from either source in isolation rarely aligns with pharmacist review \citep{baltrusaitis2019multimodal,arevalo2017gmu}. % \tm{this paragraph is not very smooth. EHRs should be introduced after text only method. And we can say, EHR data, as a rich resource to records drug side effects, have been used to screen DDIs before. However, the signals mined from EHRs are noisy, temporal and cohort dependent.}
The result is a practical gap: systems either ignore patient context (good interpolation, poor extrapolation) or overfit cohort noise (weak mechanism specificity), producing alerts that are imprecise or inapplicable bedside. 
% \tm{this paragraph is not very smooth. EHRs should be introduced after text only method. And we can say, EHR data, as a rich resource to records drug side effects, have been used to screen DDIs before. However, the signals mined from EHRs are noisy, temporal and cohort dependent.}

We introduce, to our knowledge, the first DDI framework that injects EHR context into knowledge-graph relation scoring and, within the same system, distills KG structure into an EHR-only model for zero-shot use, supporting interpolation on seen drugs and zero-shot extrapolation on unseen drugs in one system. For drugs in the graph, a fusion model combines KG and EHR features, modulating mechanism scores \citep{arevalo2017gmu,perez2018film,srivastava2015highway}; for new or rarely used drugs, a distilled EHR-only path reproduces the teacher’s mechanism behavior without KG  \citep{hinton2015distillation,lopezpaz2015generalized}, where EHR drug embeddings are derived from patient-level RETAIN and aggregated per drug before pairwise feature construction (Appendix~\ref{app:x_ehr_groups} \&~\ref{app:ehr_from_retain}) \citep{choi2016retain}. Our system can be applied to tasks such as order verification, admission and discharge reconciliation, and formulary screening. It is designed as a single clinician-facing interface that provides clear, mechanism-specific alerts pharmacists can verify, helping to reduce unnecessary warnings and avoid generic risk scores.

\section{Methods}

\subsection{Clinical objective and framework}
In this work, our goal is to predict unsafe DDIs for unseen or newly ordered drugs that are absent from KGs but present in EHRs. The primary outcome is exact mechanism matching to DrugBank’s 86-way ontology to ensure DDI alerts are actionable. We also evaluate high-precision detection (alert vs.\ no alert) and report recall-derived metrics like F1.

KG-only methods interpolate among seen drugs but falter without edges, while EHR-only methods capture bedside context but are noisy and cohort-dependent. We propose a KG–EHR teacher–student framework: a Fusion Module mixes KG and aggregated per-drug EHR embeddings derived from RETAIN \citep{choi2016retain}~(Appendix \ref{app:ehr_from_retain}) on overlapped drugs, and a distilled EHR-only student, an MLP on the same pairwise EHR features, handles drugs without KG access while preserving mechanism behavior. Thus the EHR-based student operates on drugs absent from KGs while benefiting from KG knowledge via distillation. We evaluate under two regimes (see Experimental Splits): edge hold-out, where the drug set is fixed and a fraction of edges from KGs is withheld for test while KG embeddings for all drugs remain available; and node hold-out, where a disjoint set of drugs and their incident edges are withheld during training so any test pair involving a held-out drug is evaluated zero-shot to the KG (no usable KG embedding, but may have an EHR embedding). A high-precision detection gate governs whether to alert; on alerts, the system returns mechanism-level flags for pharmacist review.

\subsection{Fusion teacher (embedding-level \& clinically motivated)}
KG and EHR entity representations are fused by a dimension-wise gate before the KG scoring model. 
Let $e_h,e_t\in\mathbb{R}^{d}$ be KG entity embeddings and $v_h,v_t\in\mathbb{R}^{d}$ the per-drug EHR embeddings. 
For $i\in\{h,t\}$,
\small
\begin{align}
\hat e_i &= P_k e_i,\quad \hat v_i = P_e v_i,\quad 
g_i &= \sigma\!\big(W_2\,\rho(W_1[\hat e_i \,\|\, \hat v_i])\big), \label{eq:gate1}\\
\tilde e_i &= g_i \odot \hat e_i + (1-g_i)\odot \hat v_i. \label{eq:gate2}
\end{align}
\normalsize
The KG scoring model then scores each relation $r$ from the fused entities,
\small
\begin{align}
s_r(h,t) &= \phi(\tilde e_h,\, e_r,\, \tilde e_t), \qquad 
r = 1,\dots,R\ (R{=}86) \label{eq:scoring model}
\end{align}
\normalsize

where $e_r$ is the (unchanged) relation embedding and $\phi$ is the AutoSF or R-GNN scoring model. 
We train with single-label cross-entropy over the $R$ logits $s=\{s_r\}_{r=1}^{R}$.\\

\subsection{Knowledge-distilled student (zero-shot)}
% Local tightening for this specific figure (affects space above AND below)
\begingroup
\setlength{\intextsep}{6pt}        % space above/below here-placed floats
\setlength{\textfloatsep}{1pt}     % space above/below top/bottom floats
\setlength{\abovecaptionskip}{1pt} % caption ↔ figure gap (caption is above with \floatconts)
\setlength{\belowcaptionskip}{1pt}

\begin{figure}[htbp]
\floatconts
  {fig:workflow2}
  {\caption{Training pipeline for the distillation model}}
  {\includegraphics[width=0.95\linewidth]{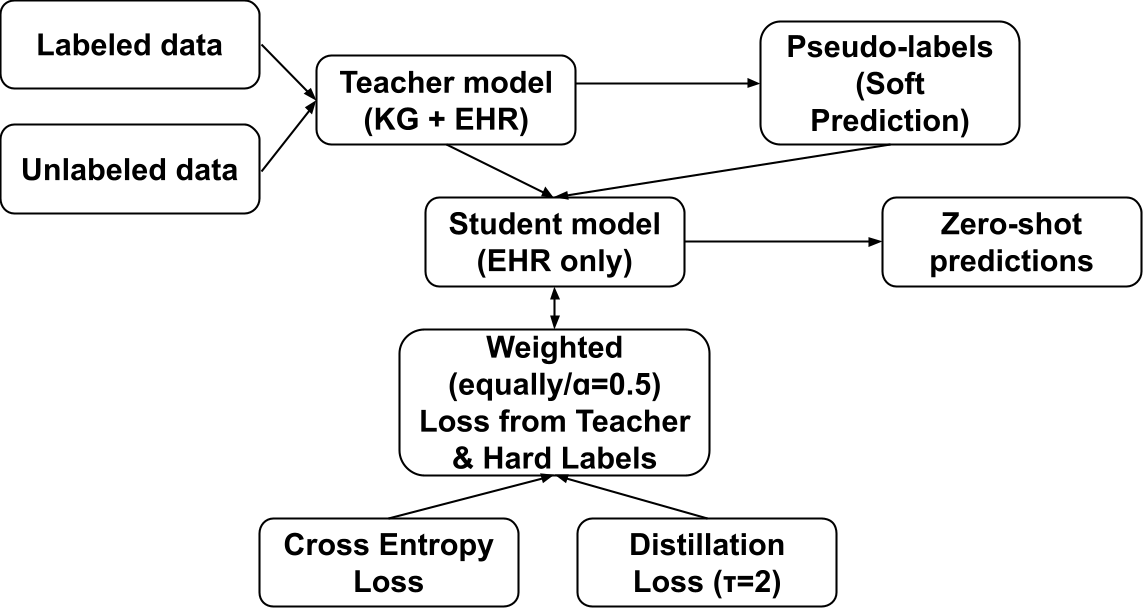}}
\end{figure}
\endgroup

The student only uses EHR feature embeddings and predicts over the same $R$ relations. The KG head exists only in the teacher. We use two heads in code: a DistMult-style head when entity embeddings are available, and an EHR head for pure EHR features:
\raggedbottom
\begin{align}
z_s &= W_2 \,\rho(W_1 x + b_1) + b_2 \in \mathbb{R}^R. \label{eq:student}
\end{align}

Distillation uses teacher logits $z_t$ converted to soft targets via a teacher-only temperature; in all experiments $\tau{=}1$ so $q=\sigma(z_t)$. Training minimizes a mixture of KD and supervised losses using BCE-with-logits on the student logits with no temperature:
% requires \usepackage{graphicx}
\begin{equation}
\resizebox{\linewidth}{!}{$
\begin{aligned}
\mathcal{L}_{\mathrm{KD}}   &= \frac{1}{R}\sum_{r=1}^{R}\!\Big[-q_r\log\sigma([\!z_s\!]_r)-(1-q_r)\log\big(1-\sigma([\!z_s\!]_r)\big)\Big],\\
\mathcal{L}_{\mathrm{sup}}  &= \frac{1}{R}\sum_{r=1}^{R}\!\Big[-y_r\log\sigma([\!z_s\!]_r)-(1-y_r)\log\big(1-\sigma([\!z_s\!]_r)\big)\Big],\\
\mathcal{L}                  &= \alpha\,\mathcal{L}_{\mathrm{KD}} + (1-\alpha)\,\mathcal{L}_{\mathrm{sup}},\quad \alpha{=}0.5,\ \tau{=}1.
\end{aligned}
$}
\label{eq:kd}
\end{equation}

If a sample lacks a hard label, the loss reduces to $\alpha\,\mathcal{L}_{\mathrm{KD}}$ \citep{hinton2015distillation,lopezpaz2015generalized}. A detailed rationale for using distillation in this setting is provided in Appendix~\ref{app:RationaleDistillation}. Further implementation details (optimizer, batch size, training epochs) are provided in 
Appendix~\ref{app:implementation}.

\section{Experiments}

\subsection{Experimental splits}
\paragraph{Transductive (edge hold-out)} We hold out 10\% edges for test and use 80\% for training and then evaluate exact-match relation ID on the held-out edges. This setting acts as a secondary check of the fusion Teacher's effectiveness.

\paragraph{Zero-shot (node hold-out)}
To assess generalization to unseen drugs, we use a node-level split: 80\% of drug nodes for training and 10\% for test; test drugs and all incident edges are removed so they are unseen at train time. Before distilling the student, we verified no leakage by remapping indices to the overlap space, building teacher-train and KD-on-train edge arrays, deduplicating and range-checking, and asserting that both endpoints of every KD edge lie in the training-node set; both checks returned OK.

We minimize per-mechanism BCE to match the teacher’s unnormalized confidences across 86 mechanisms; we found KL on softmaxed logits suppressed rare-mechanism recall.

\subsection{Detection, metrics, \& Baselines}
\lead{Two-head inference} We use detection$\rightarrow$ classification for both regimes: a binary detector decides alert vs no-alert; on alerts, the student predicts the mechanism class.

\noindent\textbf{Negatives} Detection uses leakage-safe random negatives (k=2 per positive on train, k=10 per positive on test), excluding any pair positive in any split. For detection we fix per-fold negative pools (k=10), use a single set of shared hyperparameters, and enforce leakage-safe sampling (no negatives touching test drugs).

\noindent\textbf{Reported metrics} We report exact-mechanism precision (95\% Wilson CIs) and F1 (same splits/negs).

\noindent\textbf{Baselines} We compare to KG-only (AutoSF), EHR-only, SMILES-only, and a late-fusion variant. Main metrics are precision and F1 on seen/unseen splits (Table 1).\\
\emph{AutoSF or R-GNN (KG only)} We used the AutoSF \citep{zhang2019autosf} or R-GNN \citep{10.1007/978-3-319-93417-4_38} for KG embedding and link prediction as our ablation model and baseline.\\
\emph{EHR MLP} This is the model using MLP on concatenated per-drug EHR vectors for $(h,t)$ without distillation. We use the same splits as \citep{yang2023pyhealth}.\\
\emph{SMILES MLP} Using chemical structures has also been a common method for predicting drug-drug interactions and it is also generalizable to unseen drugs, so we take SMILES as input and use ChemBERTa to obtain per-drug embeddings for DDI prediction. \citep{chithrananda2020chembertalargescaleselfsupervisedpretraining}.

\section{Results}
\label{sec:results}

\paragraph{Overview}
We report exact–mechanism precision with 95\% Wilson CIs and binary F1 with constant splits and negatives. Results are presented for two evaluation regimes: edge hold-out (seen drugs, unseen edges) and node hold-out (unseen drugs). Additional formulas for calculations in this section are provided in Appendix~\ref{app:results-details}.

% Local tightening for this table only
\begingroup
\setlength{\intextsep}{-1pt}
\setlength{\textfloatsep}{6pt}
\setlength{\abovecaptionskip}{6pt}
\setlength{\belowcaptionskip}{6pt}

\begin{table}[H]
  \centering
  \caption{Edge hold-out (seen drugs, unseen edges): precision with 95\% CIs.}
  \label{tab:edge-holdout}
  \resizebox{\linewidth}{!}{%
  \begin{tabular}{lcccc}
    \toprule
    Pipeline & + Precision & Margin of Error & F1\\
    \midrule
    \textbf{Fusion Module}    & \textbf{0.9008} & \textbf{0.0053} & \textbf{0.3198}\\
    AutoSF KG        & 0.8133 & 0.0069 & 0.1682\\
    EHR MLP          & 0.5712 & 0.0087 & 0.1718\\
    SMILES MLP       & 0.4541 & 0.0088 & 0.1667\\
    \bottomrule
  \end{tabular}}
\end{table}
\endgroup

\paragraph{Edge hold-out (seen drugs)}
The two-head Fusion (detector→mechanism) attains \textbf{0.9008} precision (95\% CI \textbf{0.895–0.906}), an \textbf{11\%} point gain over AutoSF KG at \textbf{0.8133} (95\% CI \textbf{0.806–0.820}) and far above unimodal EHR (\textbf{0.5712}) and SMILES (\textbf{0.4541}). F1 rises to \textbf{0.3198}, nearly doubling AutoSF (\textbf{0.1682}) and EHR (\textbf{0.1718}). A detection head gates edges before the mechanism classifier fires, Fusion filters away the bulk of false positives that inflate graph-only and unimodal baselines. The non-overlapping CIs establish a clear, statistically robust advantage for seen drugs.

% Local tightening for this table only
\begingroup
\setlength{\intextsep}{-6pt}
\setlength{\textfloatsep}{2pt}
\setlength{\abovecaptionskip}{2pt}
\setlength{\belowcaptionskip}{2pt}
\setlength{\dbltextfloatsep}{2pt}

\begin{table*}[t]
\centering
\caption{\textbf{est k=10 negatives per positive; same splits/negatives across models.}}
\label{tab:k10_same_splits}
\resizebox{\textwidth}{!}{%
\begin{tabular}{l c c c c c}
\hline
\textbf{Model} & \textbf{ROC-AUC} & \textbf{AP (stepwise)} & \textbf{AP (baseline)} & \textbf{F1 (detection)} & \textbf{Precision} \\
\hline
\begin{tabular}[c]{@{}l@{}}
\textbf{Distillation (teacher: Fusion head;}\\
\textbf{student distilled from R-GNN)}
\end{tabular} & $0.6223 \pm 0.0009$ & $0.2172 \pm 0.0011$ & $0.0506$ & $0.3545 \pm 0.0014$ & $0.7126 \pm 0.0345$ \\
EHR-only MLP & $0.5114 \pm 0.0008$ & $0.1697 \pm 0.0003$ & $0.0506$ & $0.2798 \pm 0.0006$ & $0.5869 \pm 0.0238$ \\
SMILES-only MLP & $0.5408 \pm 0.0094$ & $0.1794 \pm 0.0028$ & $0.0506$ & $0.2934 \pm 0.0042$ & $0.4255 \pm 0.0192$ \\
Raw R-GCN (KG-only) & $0.5176 \pm 0.0240$ & $0.1724 \pm 0.0078$ & $0.0506$ & $0.1526 \pm 0.0936$ & $0.1772 \pm 0.0111$ \\
\hline
\end{tabular}}
\end{table*}
\endgroup

\paragraph{Node hold-out} We highlight three contrasts that matter in practice. \textit{Precision:} the \textbf{Fusion Student} reaches $0.713\pm 0.035$ versus \textbf{SMILES-only} $0.426\pm 0.019$ \big($+0.287$\big) and \textbf{EHR-only} $0.587\pm 0.024$ \big($+0.126$\big), cutting the false-alert fraction at the same alert volume by roughly half vs.\ SMILES \big($1-0.713$ vs.\ $1-0.426$\big) and by $\sim 30\%$ vs.\ EHR; the $\pm$ bands do not overlap.
\textit{F1:} the Student attains $0.355$ vs.\ \textbf{EHR-only} $0.280$ \big($+0.075$\big) and \textbf{SMILES-only} $0.293$ \big($+0.062$\big), so higher precision at the shared threshold is accompanied by stronger detection quality.
\textit{Stability/cold-start:} the \textbf{KG-only R-GCN} drops to $0.177\pm 0.011$ precision with highly variable F1 $(0.153\pm 0.094)$ under $k=10$ node hold-out. Threshold-free corroboration: ROC-AUC is $0.622\pm 0.001$ (\textbf{EHR-only} $0.511\,\pm 0.001$, \textbf{SMILES-only} $0.541\,\pm 0.009$) and AP (stepwise) is $0.217\pm 0.001$ (\textbf{EHR-only} $0.170\,\pm 0.000$, \textbf{SMILES-only} $0.179\,\pm 0.003$), both well above the prevalence baseline $\sim 0.051$, meaning earlier retrieval of true interactions at a fixed review budget.
\paragraph{Case Studies}
\label{app:case-studies}
Mechanism-level case studies for Dapagliflozin and Tamoxifen evaluate pairs whose partner drugs were absent from DrugBank KG at train time. EHR-only Distilled Student flagged these interactions and assigned mechanisms consistent with clinical evidence (e.g., CYP2D6-mediated reduction of endoxifen; pharmacodynamic volume-depletion with diuretics), providing external \vspace{-2pt} signal beyond DrugBank and complementing the aggregate metrics.
\vspace{-6pt}
% Minimal preamble needs: \usepackage{graphicx}
\begin{table}[H]
\centering
\small
% phantom "table": just the image, no tabular
\caption{External signal beyond DrugBank: mechanism-level case studies.}
{\raggedleft\includegraphics[width=\linewidth]{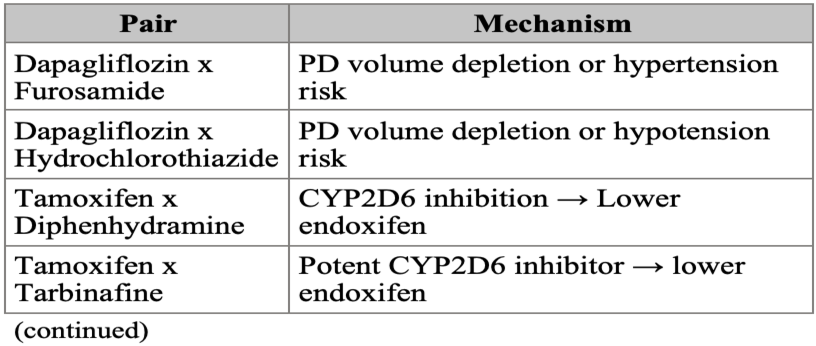}\par}
\label{tab:external-signal-case-studies}
\end{table}\raggedleft
\vspace{-12pt}
\includegraphics[width=\linewidth]{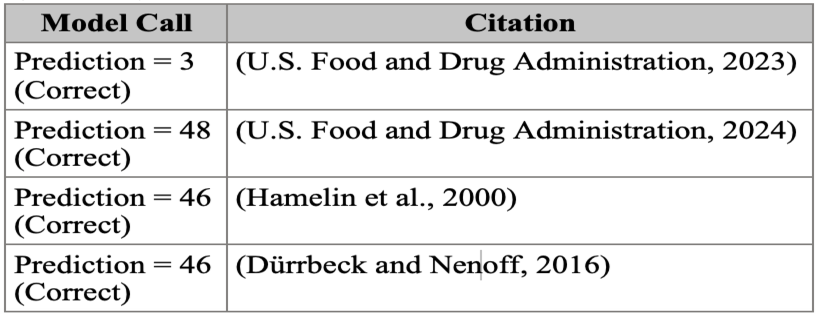}
\justifying
Despite EHRs specialized in kidney disease, the distilled Student still identifies non-renal pharmacokinetic pairs (tamoxifen), suggesting the learned pattern transfers beyond the training domain.
\paragraph{Cross-setting ablations}
Experiments and baselines act as an ablation of information sources. KG topology interpolates effectively among seen drugs (KG on edge hold-out) but fails to extrapolate to unseen nodes (node hold-out collapse). EHR features generalize to unseen drugs but are weaker than Fusion on seen drugs. Fusion integrates both, cutting false alerts while raising mechanism precision in the known-drug setting, and the KD-trained Student transfers this calibration into an EHR-only model that boosts zero-shot precision and F1 without KG access at inference. Together, these findings demonstrate that only structured integration yields consistently reliable performance across both regimes.

\paragraph{Robustness}
All key gains are supported by 95\% and bootstrap Wilson CIs with no interval overlap (Fusion vs KG; Student vs EHR). Across 3 random seeds for the zero-shot ``unseen" setting (and 1 for the transductive ``seen" setting), improvements hold across both mechanism precision and detection F1, underscoring that the observed effects reflect true signal rather than sampling noise. \emph{Margins of Error:} For precision, 95\% margins of error from Table~2 are: Fusion $0.0345$, EHR-only MLP $0.0238$, SMILES-only MLP $0.0192$, Raw R\mbox{-}GCN (KG-only) $0.0111$. \emph{Detection protocol:} For F1, we fix per-fold negative pools ($k{=}10$), use a shared threshold, and enforce leakage-safe sampling (no negatives that touch test drugs). \emph{Summary by regime: } At $\theta^\ast$, F1 is: Fusion $0.3545\pm0.0014$, EHR-only MLP $0.2798\pm0.0006$, SMILES-only MLP $0.2934\pm0.0042$, Raw R\mbox{-}GCN (KG-only) $0.1526\pm0.0936$; the ordering Fusion $>$ SMILES $>$ EHR $>$ Raw R\mbox{-}GCN holds in both regimes (edge hold-out, node hold-out).

\paragraph{Seeds} Experiments use $S{=}3$ seeds for the unseen and $S{=}1$ for seen; precision and F1 are Wilson intervals and bootstrap CIs from pooled predictions.

\section{Discussion and Outlook}
\label{sec:discussion}

\paragraph{Clinical \& algorithmic takeaways}
\textbf{Fusion }A naïve EHR+SMILES concatenation (Appendix~\ref{app:ehr-smiles}) baseline underperformed unimodal models, underscoring that simple feature-level merging cannot reconcile misaligned statistical structure and pharmacological semantics (Appendix~\ref{app:ehr-smiles}). Thus, we take the strong graph model for known drugs and add patient checks: EHR signals (e.g., co-medications, labs, QTc) amplify mechanisms they support and dampen those they contradict. This preserves relational structure while filtering patient-implausible links, resulting in higher precision, narrower 95\% confidence intervals, and lower false-positive rate.

\noindent\textbf{EHR-only Student} For new or rarely used drugs, the graph offers no help. We train an EHR-only model to mimic the teacher (distillation, $\tau{=}1$), learning its mechanism pattern and applying it with routine EHR features—no KG at runtime. The Student generalizes to unseen drugs with higher precision and better F1 than a plain EHR model: decisions are mechanism-aligned, and stable even when DrugBank KG embeddings are missing or stale.
\paragraph{Future bedside implications \& applications}
The system emits mechanism-specific, patient-conditioned flags at workload-aligned precision; Fusion covers seen formulary drugs and the Student provides zero-shot coverage for new or rare drugs (deployment details and service metrics in Appendix~\ref{app:deployment}).
\section{Conclusion}
We target precise, auditable DDI flags for pharmacist review. Prior work has largely treated KGs and EHRs in isolation; here we couple KG relation scoring with patient context and distill the structure into an EHR-only model for zero-shot use, producing the same 86-mechanism ontology. Trained and evaluated on a large EHR corpus paired with a curated DDI KG under leakage-safe edge and node hold-outs with $k=2$ negatives for training and $k=10$ negatives for testing, the system is clinically impactful, statistically sound, and deployable.

Precision and F1 improve while false positives decrease relative to AutoSF, EHR-MLP, and SMILES-MLP. Fusion raises precision and reduces bedside false alerts for seen drugs, while the student surpasses EHR-only and SMILES on unseen drugs without KG access at inference. Outputs are mechanism-aware, with robust estimates (non-overlapping CIs, tight margins, stable F1), yielding more precise alerts and lower clinical burden through patient-conditioned interpolation and distilled extrapolation.

Future work focuses on a silent-mode hospital deployment to assess precision and actionability on unseen drugs. Looking ahead, we aim to calibrate uncertainty and add abstention mechanisms to improve reliability, refresh KG structure and enable lightweight model editing over time, and conduct external and temporal validation. We will also continue to modernize the models used, scaling our pipeline with SAIL and GrAIL, and optimize hyperparameters like distillation temperature ($\tau$) through more ablation studies. These extensions will enhance generalizability and strengthen readiness for real-world deployment. To align the system with bedside use, we will operate at a clinically chosen threshold (targeting $\mathrm{TPR}\ge 0.90$ and maximizing precision on a validation cohort) while routing lower-confidence candidates to non-interruptive queues; surface interpretable evidence—the predicted mechanism and dimension-wise fusion gate weights (EHR vs.\ KG) with links to sources; support adoption via versioned models and a CPOE feedback loop (``useful'' / ``not useful'' / ``missed''); and harden safety rails with abstain/OOD flags for unfamiliar drugs and a ``silent-mode'' logging phase before any interruptive alerts.

\clearpage

\bibliography{jmlr-sample}

\clearpage

%==========================
% Appendix: EHR Feature Groups & Construction
%==========================

\appendix

\section{Information about TriNetX \citep{trinetx}}
\label{app:trinetx}

\paragraph{What it is.} TriNetX is a federated network of de-identified electronic health records (EHRs) used for cohort discovery and real-world evidence studies, described in \emph{JAMIA Open} as a global commercial network linking healthcare organizations (HCOs) with industry and academia.

\paragraph{Scale (as of 2025).} TriNetX reports 300M+ patients across 200+ HCOs and 8{,}800+ sites worldwide (press release, Sept.\ 29, 2025). An earlier update (Jan.\ 2025) cited 283M patients, and the peer-reviewed history documents growth from \textbf{55 }HCOs / 7 countries (2017) to 220+ HCOs / 30 countries (2022)—counts vary as sites onboard.

\paragraph{Data domains \& standards.} Contributed EHRs are harmonized to common terminologies (e.g., ICD-10-CM, ICD-10-PCS/HCPCS/CPT, RxNorm, LOINC), with cross-region mappings described in TriNetX terminology materials.

\paragraph{Privacy \& governance.} Data shared to researchers are de-identified under HIPAA Expert Determination. Protections include a minimum cell size of 10 for aggregates and obfuscations for person-level extracts (e.g., month–year for death, restricted geography), per an external assessment by Bradley Malin, PhD \cite{malin2020trinetx}.

\paragraph{U.S. footprint \& refresh.} Public materials note \textbf{\textasciitilde}117M U.S. patients, with datasets refreshed about every 2–4 weeks (offerings vary by product/network).

\paragraph{How we used it.} We queried the de-identified Research Network to construct cohorts and features using standard codes; no identified data left TriNetX. Many institutions treat such analyses as IRB-exempt, but local policy governs.

% Note: Replace the citation keys above with your paper's actual BibTeX keys.

\section{EHR Feature Groups Derived from TriNetX}
\label{app:x_ehr_groups}
We derive per-drug vectors from \texttt{PyHealth} \textsc{RETAIN} (exported \texttt{model.embeddings["drugs"].weight}) and augment them with structured EHR features computed pairwise for candidate drug pairs \((h,t)\). Table~\ref{tab:x_ehr_groups_transparency} lists the source tables and the pairwise features we construct. These features are concatenated to form $\mathbf{x}_{\text{EHR}}(h,t)$ for the EHR-only baseline and as inputs to the fusion gate.

\label{app:data_transparency_ehr}
We list the EHR sources used to derive per-drug EHR embeddings/features. This table is for transparency only; no pairwise feature definitions here.

\begin{table}[H]
\centering
\caption{EHR feature groups and source tables (codes).}
\label{tab:x_ehr_groups_transparency}
\resizebox{\linewidth}{!}{%
\begin{tabular}{p{0.28\linewidth} p{0.68\linewidth}}
\toprule
\textbf{Group} & \textbf{Source tables (codes)} \\
\midrule
Demographics & \texttt{patient.csv} \\
Encounters & \texttt{encounter.csv} \\
Diagnoses & \texttt{diagnosis.csv} (ICD-10) \\
Meds (orders/dispense) & \texttt{medication\_drug.csv} (RxNorm/NDC) \\
Meds (ingredients) & \texttt{medication\_ingredient.csv} \\
Procedures & \texttt{procedure.csv} (CPT) \\
Labs & \texttt{lab\_result.csv} (LOINC) \\
Vitals & \texttt{vital\_signs.csv} \\
\bottomrule
\end{tabular}}
\end{table}

\section{Deriving EHR Drug Embeddings from \texttt{PyHealth} \textsc{RETAIN}}
\label{app:ehr_from_retain}

\noindent We train \textsc{RETAIN} with \texttt{PyHealth} using \texttt{feature\_keys=["drugs","procedures"]} and a multi-label objective. During training, \textsc{RETAIN} learns a global token embedding table for the \texttt{"drugs"} stream:
\[
W \in \mathbb{R}^{V \times D}, \qquad D=64 \text{ (our setting)}.
\]
After training, we export this layer exactly as in code:
\[
\texttt{emb\_layer} := \texttt{model.embeddings["drugs"]}\]
\[
W := \texttt{emb\_layer.weight}.
\]

Let \(\text{id}(i)\) be the PyHealth vocabulary index for drug \(i\) (built from the training split). The \emph{EHR embedding} we use for drug \(i\) is the corresponding row of \(W\):
\[
\mathbf{v}_i \;=\; W_{\text{id}(i)} \in \mathbb{R}^{D}.
\]

Downstream pairwise EHR features for a candidate pair \((h,t)\) are constructed from these exported vectors:
\[
\mathbf{x}_{\text{EHR}}(h,t) \;=\;
\big[\; \mathbf{v}_h \ \|\ \mathbf{v}_t \;\big],
\]
or optionally
\[
\mathbf{x}_{\text{EHR}}(h,t) \;=\;
\big[\; \mathbf{v}_h \ \|\ \mathbf{v}_t \ \|\ |\mathbf{v}_h-\mathbf{v}_t| \ \|\ \mathbf{v}_h \odot \mathbf{v}_t \;\big],
\]
after \(\ell_2\) normalization of \(\mathbf{v}_i\). These \(\mathbf{x}_{\text{EHR}}\) features are used both in the EHR-only MLP baseline and as inputs to the fusion gate.

\section{Rationale for Fusion \& Distillation}
\label{app:RationaleDistillation}
\noindent\textbf{Rationale for ChemBERTa }Why ChemBERTa (zinc-base-v1). It is a stable, off-the-shelf SMILES encoder with good coverage and straightforward integration for our ablations.

\noindent\textbf{Fusion objective: Why this gate }The gate is parameter-efficient, easy to optimize under multi-relation supervision, and interpretable: it turns up KG or EHR signal dimension-by-dimension in ways clinicians can inspect (mechanism evidence $\uparrow$, contradictory evidence $\downarrow$), while preserving KG relational geometry. Empirically, naive concatenation $+$ MLP underperforms both unimodal models, motivating an explicit gate rather than “just stacking features.”

\emph{Mathematical form already in the paper.}
\begin{align}
\hat{e}_i &= P_k e_i \\
\hat{v}_i &= P_e v_i \\
g_i &= \sigma\!\bigl(W_2\,\rho\!\bigl(W_1[\hat{e}_i;\hat{v}_i]\bigr)\bigr) \tag{8}
\end{align}

\begin{equation}
s_r(h,t) \;=\; \phi\!\bigl(\tilde{e}_h,\, e_r,\, \tilde{e}_t\bigr), 
\qquad r = 1,\dots,86 \tag{9}
\end{equation}

\medskip
\noindent\textbf{Backbone robustness (not tied to a single scorer).}
Under train/test node splits of $80\%$ and $10\%$, a PyG R\,-GCN backbone (KG-only) yields higher precision than AutoSF, while preserving the ordering \textsc{Fusion} $>$ KG-only and the Student’s zero-shot advantages:
\[
\begin{aligned}
\text{R-GCN per-seed precision: }&\ 0.7352,\, 0.7297,\, 0.6729 \\
&\to\ \text{ mean } 0.7126 \pm 0.0345
\end{aligned}
\]
\[
\text{AutoSF: } 0.6983,\, 0.6707,\, 0.6443 \;\to\; \text{mean } 0.6711 \pm 0.0270.
\]
Multi-seed results for baselines will be in the camera-ready version.
\noindent\textbf{Distillation objective }
DrugBank mechanisms are predominantly single-label, but teacher confidences over $R{=}86$ are not a normalized distribution and can place mass on multiple plausible mechanisms. BCE treats mechanisms independently and avoids the probability-mass competition imposed by $\mathrm{KL}(\mathrm{softmax}_\tau(z_t)\,\|\,\mathrm{softmax}(z_s))$, which we observed to suppress rare-mechanism recall. BCE was also more numerically stable with many negatives.

% Comparison matrix: our method vs. baselines
% (Fits in one column; uses simple check/times symbols)

\section{Method Positioning vs. Baselines}
\label{app:method_vs_baselines}

\newcommand{\cmark}{$\checkmark$}
\newcommand{\xmark}{$\times$}

% requires \usepackage{booktabs} and \usepackage{graphicx}
% (and, if not already defined:)
% \newcommand{\cmark}{$\checkmark$}
% \newcommand{\xmark}{$\times$}

\begin{table}[H]
\centering
\caption{\textbf{Model capabilities and inputs.} Our Fusion teacher combines KG + EHR with a gating module; the Student is an EHR-only model distilled from the teacher. All models use the same 86-mechanism labels and splits/negatives.}
\label{tab:method_differentiation}

% -------- Part 1: data modalities --------
\resizebox{\linewidth}{!}{%
\begin{tabular}{l c c c}
\toprule
\textbf{Method} & \textbf{Uses KG} & \textbf{Uses EHR} & \textbf{Uses SMILES} \\
\midrule
\textbf{Ours: Fusion (teacher)} & \cmark & \cmark & \xmark \\
\textbf{Ours: Student (distilled)} & \xmark & \cmark & \xmark \\
KG-only: AutoSF & \cmark & \xmark & \xmark \\
KG-only: R-GCN & \cmark & \xmark & \xmark \\
EHR-only: MLP (RETAIN $W$) & \xmark & \cmark & \xmark \\
SMILES-only: MLP & \xmark & \xmark & \cmark \\
\bottomrule
\end{tabular}
}

\vspace{4pt}

% -------- Part 2: mechanisms & training scheme --------
\resizebox{\linewidth}{!}{%
\begin{tabular}{l c c c c}
\toprule
\textbf{Method} & \textbf{Fusion gate} & \textbf{KD (teacher$\rightarrow$student)} & \textbf{86 mech.} & \textbf{Zero-shot drugs} \\
\midrule
\textbf{Ours: Fusion (teacher)} & \cmark & \xmark & \cmark & \cmark \\
\textbf{Ours: Student (distilled)} & \xmark & \cmark & \cmark & \cmark \\
KG-only: AutoSF & \xmark & \xmark & \cmark & \xmark \\
KG-only: R-GCN & \xmark & \xmark & \cmark & \xmark \\
EHR-only: MLP (RETAIN $W$) & \xmark & \xmark & \cmark & \cmark \\
SMILES-only: MLP & \xmark & \xmark & \cmark & \xmark \\
\bottomrule
\end{tabular}
}
\end{table}

\paragraph{Notes.}
“Uses EHR” denotes structured/EHR-derived per-drug vectors (exported from \texttt{PyHealth} \textsc{RETAIN} token embedding table $W$) and pairwise features; “Zero-shot drugs” indicates the model can score pairs where both drugs are unseen in the KG (enabled via EHR inputs for our teacher/student).

% Requires: \usepackage{booktabs,tabularx}
% Requires in your preamble:
% \usepackage{booktabs,tabularx,array}
% \newcolumntype{Y}{>{\raggedright\arraybackslash}X}

% Preamble needs:
% \usepackage{booktabs,array}
% \newcolumntype{P}[1]{>{\raggedright\arraybackslash}p{#1}}

% Preamble needs:
% \usepackage{booktabs,array,graphicx}
% \newcolumntype{P}[1]{>{\raggedright\arraybackslash}p{#1}}

% Preamble needs:
% \usepackage{booktabs,array,graphicx,makecell}
% \newcolumntype{P}[1]{>{\raggedright\arraybackslash}p{#1}}

% Preamble:
% \usepackage{booktabs,array}
% \newcolumntype{P}[1]{>{\raggedright\arraybackslash}p{#1}}

% Preamble:
% \usepackage{booktabs,array}
% \newcolumntype{P}[1]{>{\raggedright\arraybackslash}p{#1}}
% \tm{this is good, but I am wondering if we need to include it in the main paper. We can just say in the discussion section: we tried to combine SMILES+EHR, and it is worse than using EHR only, so we only take EHR MLP as the student.}

\section{Additional Baseline: EHR+SMILES}
\label{app:ehr-smiles}

\paragraph{Motivation.} 
We additionally tested a naïve concatenation baseline combining per-drug EHR embeddings with ChemBERTa-derived SMILES embeddings, followed by an MLP classifier. This baseline was included to assess whether molecular structure embeddings could enhance patient-derived features without KG support. 

\paragraph{Results.} 
The EHR+SMILES baseline underperformed both unimodal models. In edge hold-out (seen drugs, unseen edges), it achieved a precision of $0.2818$, below both EHR and SMILES-only. In node hold-out (unseen drugs), it yielded $0.3027$ with $F1{=}0$, indicating that the model admitted no alerts. These results highlight the difficulty of integrating heterogeneous embeddings whose statistical structure and pharmacological semantics are misaligned, reinforcing that simple feature-level merging does not substitute for KG-grounded relational alignment. Statistics, appended to existing data, are shown in Tables~\ref{tab:edge-holdout-EHR-SMILES} and \ref{tab:k10_same_splits_ehr_smiles}.

% Local tightening for these tables only
\begingroup
\setlength{\intextsep}{-1pt}
\setlength{\textfloatsep}{6pt}
\setlength{\abovecaptionskip}{6pt}
\setlength{\belowcaptionskip}{6pt}

\begin{table}[H]
  \centering
  \caption{Edge hold-out (seen drugs, unseen edges): precision with 95\% CIs.}
  \label{tab:edge-holdout-EHR-SMILES}
  \resizebox{\linewidth}{!}{%
  \begin{tabular}{lcccc}
    \toprule
    Pipeline & + Precision & Margin of Error & F1\\
    \midrule
    \textbf{Fusion Module}    & \textbf{0.9008} & \textbf{0.0053} & \textbf{0.3198}\\
    AutoSF KG        & 0.8133 & 0.0069 & 0.1682\\
    EHR MLP          & 0.5712 & 0.0087 & 0.1718\\
    SMILES MLP       & 0.4541 & 0.0088 & 0.1667\\
    EHR+SMILES MLP   & 0.2818 & 0.0056 & —\\
    \bottomrule
  \end{tabular}}
\end{table}
\endgroup

\begingroup
\setlength{\intextsep}{-6pt}
\setlength{\textfloatsep}{2pt}
\setlength{\abovecaptionskip}{2pt}
\setlength{\belowcaptionskip}{2pt}

\begin{table*}[t]
\centering
\caption{\textbf{k=10 test on the same splits/negatives for all models—ROC-AUC, AP (stepwise), AP (baseline), F1, and precision.}}
\label{tab:k10_same_splits_ehr_smiles}
\resizebox{\textwidth}{!}{%
\begin{tabular}{l c c c c c}
\hline
\textbf{Model} & \textbf{ROC-AUC} & \textbf{AP (stepwise)} & \textbf{AP (baseline)} & \textbf{F1 (detection)} & \textbf{Precision} \\
\hline
\begin{tabular}[c]{@{}l@{}}
\textbf{Distillation (teacher: Fusion head;}\\
\textbf{student distilled from R-GNN)}
\end{tabular} & $0.6223 \pm 0.0009$ & $0.2172 \pm 0.0011$ & $0.0506$ & $0.3545 \pm 0.0014$ & $0.7126 \pm 0.0345$ \\
EHR-only MLP & $0.5114 \pm 0.0008$ & $0.1697 \pm 0.0003$ & $0.0506$ & $0.2798 \pm 0.0006$ & $0.5869 \pm 0.0238$ \\
SMILES-only MLP & $0.5408 \pm 0.0094$ & $0.1794 \pm 0.0028$ & $0.0506$ & $0.2934 \pm 0.0042$ & $0.4255 \pm 0.0192$ \\
EHR+SMILES MLP & — & — & — & $0.0000$ & $0.3027\pm 0.0116$ \\
Raw R-GCN (KG-only) & $0.5176 \pm 0.0240$ & $0.1724 \pm 0.0078$ & $0.0506$ & $0.1526 \pm 0.0936$ & $0.1772 \pm 0.0111$ \\
\hline
\end{tabular}}
\end{table*}
\endgroup

\section{Implementation \& Training}
\label{app:implementation}

\subsection{KG-only Hyperparameters}
\label{app:autosf}
\begin{verbatim}
python train.py --task_dir=KG_Data/DrugBank \
  --optim=Adam --lamb=0.000282 --lr=0.3775 \
  --n_dim=64 --n_epoch=250 --n_batch=2048 \
  --epoch_per_test=250 --test_batch_size=50 \
  --thres=0.0 --parrel=1 --decay_rate=0.99456
\end{verbatim}

\subsection{Fusion Module Optimization (R-GNN)}
\label{app:fusion}
\begin{itemize}
  \item Learning rate: \textbf{0.001}
  \item Training epochs: \textbf{10}
\end{itemize}

\subsection{EHR MLP (scikit-learn)}
\label{app:mlp}
\begin{verbatim}
from sklearn.neural_network \
import MLPClassifier

clf = MLPClassifier(
    alpha=0.05,
    hidden_layer_sizes=(32, 16, 8, 4, 2),
    max_iter=30,
    random_state=1
)
\end{verbatim}

\paragraph{Hyperparameters}
\begin{enumerate}
    \item alpha: learning rate
    \item hidden\_layer\_sizes: self-explanatory
    \item max\_iter: number of epochs
    \item random\_state: random seed (changed for multi-seed experiments)
\end{enumerate}

\subsection{SMILES MLP (scikit-learn)}
\label{app:mlp}
\begin{verbatim}
from sklearn.neural_network \
import MLPClassifier

clf = MLPClassifier(
    alpha=0.10,
    hidden_layer_sizes=(32, 16, 8, 4, 2),
    max_iter=15,
    random_state=1
)
\end{verbatim}

\subsection{EHR+SMILES MLP (scikit-learn)}
\label{app:mlp}
\begin{verbatim}
from sklearn.neural_network \
import MLPClassifier

clf = MLPClassifier(
    alpha=0.05,
    hidden_layer_sizes=(32, 16, 8, 4, 2),
    max_iter=50,
    random_state=1
)
\end{verbatim}

\subsection{SMILES Encoder (HuggingFace Transformers)}
\label{app:chemberta}
\begin{verbatim}
from transformers \
import AutoTokenizer, AutoModel

MODEL = "seyonec/ChemBERTa-zinc-base-v1"
tok = AutoTokenizer.from_pretrained(MODEL)
enc = AutoModel.from_pretrained(MODEL)
\end{verbatim}

\subsection{KD Student Configuration}
\label{app:kd-student}
\begin{verbatim}
KD_ALPHA   = 0.5      # $\alpha$
TAU        = 1        # temperature
LR         = 0.1
WD         = 1e-4
BATCH_SIZE = 1024
EPOCHS     = 30
n_rel      = 86
n_dim      = 64
DEVICE     = torch.device("cuda" \
if torch.cuda.is_available() else "cpu")
\end{verbatim}

\paragraph{Reproducibility note.}
All hyperparameters and commands above reflect the exact values used in the experiments reported in the main text.

\subsection{Training Time}
\begin{enumerate}
\item AutoSF (KG-only): 250 epochs on NVIDIA T4, $\approx$ 5 minutes total.

\item Relational GNN (R-GCN): 30 epochs on Colab High-RAM CPU, $\approx$ 3 minutes total.

\item Both runs use 112,008 KG triples (head–relation–tail) as training rows. Implementation commands/hyperparameters are already listed in Appendix~\ref{app:implementation}.
\end{enumerate}

\section{Results: Additional Details}
\label{app:results-details}

\paragraph{Metrics and confidence intervals.}
We report (i) exact-mechanism precision
\(
\mathrm{Prec} = \frac{1}{N}\sum_{i=1}^{N}\mathbf{1}\{\arg\max_{r\in\{1,\ldots,R\}}\hat p_i(r)=y_i\}
\)
\(
\mathrm{Prec}_{\text{bin}}=\tfrac{\mathrm{TP}}{\mathrm{TP}+\mathrm{FP}},\
\mathrm{Rec}=\tfrac{\mathrm{TP}}{\mathrm{TP}+\mathrm{FN}},\
\mathrm{F1}=\tfrac{2\,\mathrm{Prec}_{\text{bin}}\cdot \mathrm{Rec}}{\mathrm{Prec}_{\text{bin}}+\mathrm{Rec}}
\)

Wilson 95\% CI for a proportion $\hat p$ over $n$ samples:
\[
\mathrm{CI}_{\mathrm{low,high}}=
\frac{\hat p+\tfrac{z^2}{2n}\ \mp\ z\sqrt{\tfrac{\hat p(1-\hat p)}{n}+\tfrac{z^2}{4n^2}}}{1+\tfrac{z^2}{n}},
\quad z=1.96.
\]

For the edge-vs-none task we construct a test set with a ratio $k$ negatives per positive ($k{=}2$ in the main results).
A sampled pair $(h,t)$ is \emph{excluded} if it is positive in any split or direction; self-loops and inverse duplicates are removed to avoid leakage.
All models are evaluated on the same sets.

\paragraph{Relative false-positive reduction.}
At fixed true positives across methods, $\mathrm{FP}=\mathrm{TP}\big(\tfrac{1}{p}-1\big)$ for precision $p$.
The relative FP reduction of Fusion over a baseline is
\[
1-\frac{\tfrac{1}{p_{\text{fusion}}}-1}{\tfrac{1}{p_{\text{base}}}-1}.
\]

\paragraph{Seeds and uncertainty.}
Experiments were run with $S{=}3$ random seeds in the zero-shot ``unseen" setting and $S{=}1$ in the transductive ``seen" setting. For single-seed results, we report Wilson 95\% CIs for precision from pooled predictions, and F1 is reported as a point estimate. For multi-seed results, we incorporate both Wilson CIs and bootstrap CIs for each metric.

\paragraph{Tables referenced in the body.}
\textbf{Table 1 (Seen/edge hold-out):} Fusion vs AutoSF, precision with 95\% CIs.\\
\textbf{Table 2 (Unseen/node hold-out):} Student vs EHR-MLP, SMILES-MLP, AutoSF; precision with 95\% CIs and F1 (± across seeds).

% \section{Performance of SMILES+EHR MLP}
% The naive concatenation of ChemBERTa per-drug embeddings with pairwise EHR features under a single MLP underperforms both parents. The modalities differ in scale and semantics (static molecular tokens vs.\ co-prescription context), and without cross-modal calibration (e.g., per-channel normalization or a learned adapter) the MLP overfits to the sparse ChemBERTa channel. Under the high-precision operating point, the detection gate then suppresses nearly all alerts, yielding non-zero precision but $\mathrm{recall}\!\approx\!0$ and thus $\mathrm{F1}\!=\!0$ (Table~\ref{tab:node-holdout}).

\section{Deployment details}
\label{app:deployment}

For pharmacist order verification, alerts must be \emph{actionable} and \emph{rare}: the system will emit a top mechanism with a short, patient-conditioned rationale (e.g., co-medications, labs) and an audit trail; validation-fixed thresholds keep operating characteristics stable for governance \citep{shang2019gamenet,yang2021safedrug,choi2016retain}. Use cases include formulary screening, coverage for investigational or rare drugs (Student), antimicrobial stewardship and oncology checks (Fusion), and admission/discharge reconciliation when KG embeddings may be stale \citep{huang2024txgnn,wang2023zerobind,zitnik2018decagon}. Integration is EHR-native (Student uses tabular features only) and supports service metrics such as false alerts/100 orders, time-to-clear, mitigation acceptance, plus model metrics for QA and safety review \citep{pedregosa2011sklearn,yang2023pyhealth}.
\end{document}